\documentclass{llncs}
\usepackage{graphicx}
\usepackage{amssymb}
\usepackage[numbers]{natbib}
\usepackage{subfig}
\usepackage[ruled,vlined]{algorithm2e}
\usepackage{amsmath}
\usepackage{lineno}
\usepackage{changepage}
\usepackage{rotating}
\usepackage{algcompatible}
\usepackage{soul}
\usepackage[%
    colorlinks=true,
    pdfborder={0 0 0},
    linkcolor=blue
]{hyperref}
\hypersetup{
    colorlinks=true,
    linkcolor=red,
    filecolor=magenta,      
    urlcolor=blue,
    citecolor=green,
}
\usepackage{multirow}
\usepackage{svg}
\usepackage{soul}
\newcommand{\orcid}[1]{\href{https://orcid.org/#1}{\includegraphics[scale=0.7]{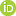}}}

\begin{document}

\title{Cervical Cytology Classification Using PCA \& GWO Enhanced Deep Features Selection}

\author{Hritam Basak \orcid{0000-0001-5921-1230}\inst{1} \and 
Rohit Kundu \orcid{0000-0001-8665-8898}\inst{1} \and
Sukanta Chakraborty\inst{2} \and
Nibaran Das \orcid{0000-0002-2426-9915}\inst{3}
}
\institute{Department of Electrical Engineering, Jadavpur University, Kolkata, India \and
Theism Medical Diagnostics Centre, Kolkata, India \and
Department of Computer Science \& Engineering, Jadavpur University, Kolkata, India
}
\authorrunning{Hritam Basak et al.}


\section*{Cervical Cytology Classification Using PCA \& GWO Enhanced Deep Features Selection}
\begin{center}
    \textbf{Hritam Basak}\\
    Department of Electrical Engineering, Jadavpur University\\
    188, Raja S.C. Mullick Road,\\
    Jadavpur, Kolkata-700032, West Bengal, INDIA\\
    Email: hritambasak48@gmail.com
\end{center}
\begin{center}
    \textbf{Rohit Kundu}\\
    Department of Electrical Engineering, Jadavpur University\\
    188, Raja S.C. Mullick Road,\\
    Jadavpur, Kolkata-700032, West Bengal, INDIA\\
    Email: rohitkunduju@gmail.com
\end{center}
\begin{center}
    \textbf{Sukanta Chakraborty}\\
    Theism Medical Diagnostics Centre\\
    Dum Dum, Kolkata-700030, West Bengal, INDIA\\
    Email: drsukantachakraborty@gmail.com
\end{center}
\begin{center}
    \textbf{Nibaran Das*}\\
    Department of Computer Science \& Engineering, Jadavpur University\\
    188, Raja S.C. Mullick Road,\\
    Jadavpur, Kolkata-700032, West Bengal, INDIA\\
    Email: nibaran.das@jadavpuruniversity.in
\end{center}
\begin{center}
    *Corresponding author: Nibaran Das\\
    *Corresponding author email: nibaran.das@jadavpuruniversity.in\\
    *Corresponding author contact number: +913324572407
\end{center}

\maketitle

\begin{abstract}
Cervical cancer is one of the most deadly and common diseases among women worldwide. It is completely curable if diagnosed in an early stage, but the tedious and costly detection procedure makes it unviable to conduct population-wise screening. Thus, to augment the effort of the clinicians, in this paper, we propose a fully automated framework that utilizes Deep Learning and feature selection using evolutionary optimization for cytology image classification. The proposed framework extracts Deep feature from several Convolution Neural Network models and uses a two-step feature reduction approach to \textcolor{black}{to ensure reduction in computation cost and faster convergence}. The features extracted from the CNN models form a large feature space whose dimensionality is reduced using Principal Component Analysis while preserving 99\% of the variance. A non-redundant, optimal feature subset is selected from this feature space using an evolutionary optimization algorithm, the Grey Wolf Optimizer, thus improving the classification performance. Finally, the selected feature subset is used to train an SVM classifier for generating the final predictions. The proposed framework is evaluated on three publicly available benchmark datasets: Mendeley Liquid Based Cytology (4-class) dataset, Herlev Pap Smear (7-class) dataset, and the SIPaKMeD Pap Smear (5-class) dataset achieving classification accuracies of 99.47\%, 98.32\% and 97.87\% respectively, thus justifying the reliability of the approach. \textcolor{black}{The relevant codes for the proposed approach can be found in: \href{https://github.com/DVLP-CMATERJU/Two-Step-Feature-Enhancement}{GitHub}}
\end{abstract}

\section{Introduction}
\begin{figure*}
\begin{adjustwidth}{-2cm}{}
    \centering
    \includegraphics[scale=0.5]{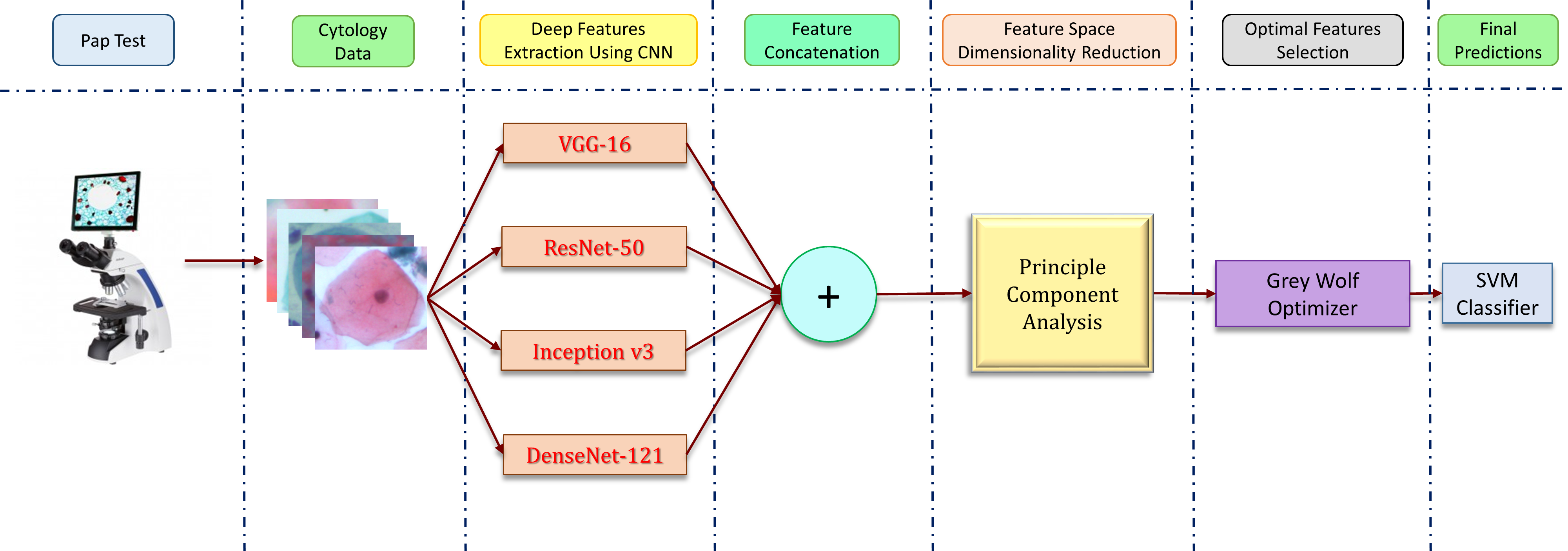}
    \caption{Overall workflow of the proposed framework}
    \label{overall}
\end{adjustwidth}
\end{figure*}
Cancer is the second leading cause of death worldwide, causing around 9.6 million deaths every year and about 1/6th of the total deaths of the population throughout the globe. Studies also suggest that the economic impact of cancer is significant as most of these deaths have been reported from poor and middle-income countries where the living index is low and healthcare infrastructure is comparatively insufficient for diagnosis and potential treatments. Among them, cervical cancer is the fourth most common cancer worldwide, having around 570 thousands of reported cases every year and the second most common cancer in women causing around 311 thousands of deaths per year \cite{ferlay2019estimating}.

Pap smear test is currently the most reliable and renowned screening tool for the detection of pre-cancerous cells and cervical lesions in women based on microscopic studies of the cells. However, the process is time-consuming since pathologists need a lot of time to classify each cell from a slide of over 10,000 cells. Thus, an automated detection tool needs to be developed to classify pre-cancerous lesions for early detection and widespread screening.

With the advent of artificial intelligence and deep learning in the domain of medical sciences and healthcare \cite{basak2020comparative, basak2021dfenet,basak2020monocular}, it is more becoming to lie on the results predicted by this decision-support system \cite{basak2020single,chattopadhyay2020multi} to undermine the observer-bias issues. In this paper, we seek to develop an alternative approach that utilizes the Deep learning-based feature extraction and optimization algorithm that gives excellent multi-class classification accuracy, performing robustly and outperforming several existing methods like \cite{gencctav2012unsupervised, win2020computer, chankong2014automatic, bora2017automated, Mitra2019,mitra2020cytology, Dey2020}.

In this research, we extracted deep features \cite{azaza2021off,10.1007/978-981-16-1086-8_4} from pre-trained CNN models and concatenated the features to form a large feature space. This is followed by the application of the Principal Component Analysis (PCA) method for dimensionality reduction of the feature space, while keeping most of the important features intact, for which we retained 99\% of the variance of the feature space. Some of the features extracted from the CNN classifiers might be non-informative or might even lead to a higher misclassification rate. To eliminate such redundant, misleading features, generally, evolutionary optimization algorithms are preferred by researchers, but applying such algorithms directly to CNN-extracted features leads to computational wastage due to the very high dimensionality of the feature space. 
\textcolor{black}{PCA reduces the dimensionality of the feature space or number of data points by combining highly correlated variables to form a smaller number of new variables, known as "principal components". From PCA, we also get higher variations in the data, even in a lower dimension. This new feature space of dimensionality lower than the original space, when used as an input to a metaheuristic optimization algorithm, significantly reduces the computation cost since a lower population size is required for faster convergence. The number of iterations required for reaching the global optima is also greatly reduced (i.e., faster evolution).}

Specifically, among different evolutionary optimization algorithms available, we used the Grey Wolf Optimizer (GWO) \cite{mirjalili2014grey} for optimal feature set selection. The GWO is embedded with a Support Vector Machines (SVM) classifier \cite{zhang2012support} with Radial Basis Function (RBF) kernel for fitness assignment and final classification. This method significantly decreased the training time for the classification task, while maintaining the competitive accuracy of predictions. The overall workflow of the proposed framework is shown in Figure \ref{overall}.

The contributions of this paper can be summarized as follows:

\begin{enumerate}
    \item In the current paper, we propose a framework for the optimal selection of deep features extracted from Convolutional Neural Network (CNN) classifiers.
    \item The dimensionality of the feature set extracted from the CNNs is large and thus Principal Component Analysis (PCA) is used to reduce the dimensionality while retaining the highly discriminating features. \textcolor{black}{The resulting feature subset, when used as the input of GWO, reduces the computation and ensures faster convergence.}
    \item Optimal features are selected through the use of a nature-inspired evolutionary optimization algorithm, the Grey Wolf Optimizer (GWO), for the first time in the cervical cytology domain, which filters out only the non-redundant features for making the final predictions.
    \item The proposed framework has been evaluated on three publicly available datasets: the Herlev Pap Smear dataset \cite{jantzen2005pap}, the SIPaKMeD Pap Smear dataset \cite{plissiti2018sipakmed} and the Mendeley Liquid Based Cytology dataset \cite{hussain2020liquid}, achieving classification accuracies of 98.32\%, 97.87\% and 99.47\% respectively. The proposed method outperformed traditional CNN based approaches and is comparable to state-of-the-art methods.
\end{enumerate}

The rest of the paper is organized as follows: Section~\ref{relwork} surveys some of the recent developments in the automated detection of cervical cancer; Section~\ref{methods} describes the proposed methodology in detail; Section~\ref{results} evaluates the performance of the proposed framework on three publicly available benchmark datasets and Section~\ref{conclusions} concludes the paper.
\section{Related Work}\label{relwork}
Previous studies show that feature extraction and selecting non-redundant features is an important part of the classification process and it affects the classification result significantly. Different methods have been explored over the years, like traditionally handcrafted feature extraction and feature selection \cite{almubarak2019hybrid}, Simulated Annealing (SA)\cite{wang2005simulated}, Convolutional Neural Networks \cite{wu2018automatic}, Fuzzy-C means \cite{william2019cervical}, to name a few. Some have given very good results in binary classification but not so much in multi-class classification, only a few have successfully given good results even for multi-class classification problem \cite{bora2017automated, chankong2014automatic, martinez2020classifying}. One of the major concerns is the availability of open-access datasets and the volume of images available in them for each class, which is a major setback for a lot of proposed methods in the literature. 

Chankong et al. \cite{chankong2014automatic} used different classifiers for binary and multi-class classification with the best accuracy obtained using ANN. William et al. \cite{william2019pap} used an enhanced Fuzzy C-means by extracting features from the cell images in Herlev Dataset and got an accuracy of 98.88\%. Byriel et al. \cite{byriel1999neuro} used the ANFIS neuro-fuzzy classification technique which achieved an accuracy of 96\% on the 2-class problem but performed much worse on the 7-class problem. Zhang et al. \cite{zhang2014automation} used MLBC slides stained in H\&E in two ways: once posing it as a 2-class dataset and again using it as a 7-class dataset. They used the Fuzzy Clustering-means approach and got an accuracy of 96-97\% in the 2-class problem and 72-77\% on the 7-class problem. Bora et al. \cite{bora2017automated} used Ensemble classifier on the Herlev Dataset and got an accuracy of 96.5\%. Marinakis et al. \cite{marinakis2009pap} in a way similar to \cite{zhang2014automation, xue2020application} of using the Herlev Dataset as both a 2-class and a 7-class problem, used genetic algorithm combined with nearest neighbour classifier and got the best result with 1 nearest neighbour classifier giving an accuracy of 98.14\% on the 2-class problem and 96.95\% on the 7-class problem both in 10-fold cross-validation. Zhang et al. \cite{zhang2017deeppap} used a deep convolutional neural network architecture thus removing the need for cell segmentation like \cite{huang2020nucleus, chattopadhyay2020multi} on the Herlev Dataset and achieved an accuracy of 98.3\%.

However, the number of publicly available datasets related to the smear images of cervical cytology is quite less and each of them contains only nearly a thousand images or less. So it becomes quite difficult to design a classical deep learning or machine learning model with that few images to classify between these images with an improved accuracy than the preexisting methods. However, transfer learning can be used for this purpose that can quite significantly tackle this issue where we use a pre-trained model (for example ResNet-50 trained with ImageNet \cite{deng2009imagenet}), fine-tune the model, and use that for the classification purpose. Akter et al. \cite{akter2021prediction} performed experimentations with different machine learning classifiers performed detailed comparative analysis on their performance. Data augmentation can be another solution where we can virtually increase the dataset size by slight movement or rotation or some other changes of the images. However, these methods cannot improve the results significantly as they cannot add more features or information to the algorithm to learn from. Therefore, as suggested by \cite{niedzielewski2020multidimensional}, we tried a new optimal feature selection approach for improving the classification accuracy and robustness of the task.
\section{Materials and Method}\label{methods}
The experiment consisted of the following steps: (1) Data acquisition (collecting image datasets of Pap smear test results from different sources), (2) Data Preprocessing (structuring the data incorrect formats and verifying the datasets), (3) Feature extraction (Extracting the important features from the datasets using different CNN models), (4a) Combining the features from different CNNs (to increase the effectiveness of the features) (4b) Feature reduction using Principal Component Analysis (PCA) method (to discard the redundant features and to improve the classification time), (5) Fitting these features to the classifier and (6) Analysis of the results. The whole task was performed using a machine having NVIDIA Tesla K80 GPU with 12 GB of available RAM size.

\subsection{Datasets Used}
\begin{figure*}
\begin{adjustwidth}{-1.5cm}{}
    \centering
    \subfloat[Herlev Dataset]{\includegraphics[scale=0.5]{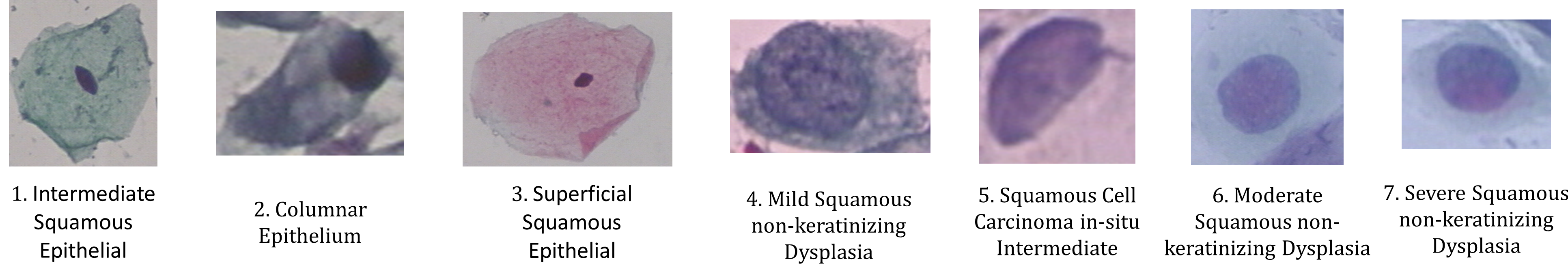}}\\
    \subfloat[Mendeley Dataset]{\includegraphics[scale=0.5]{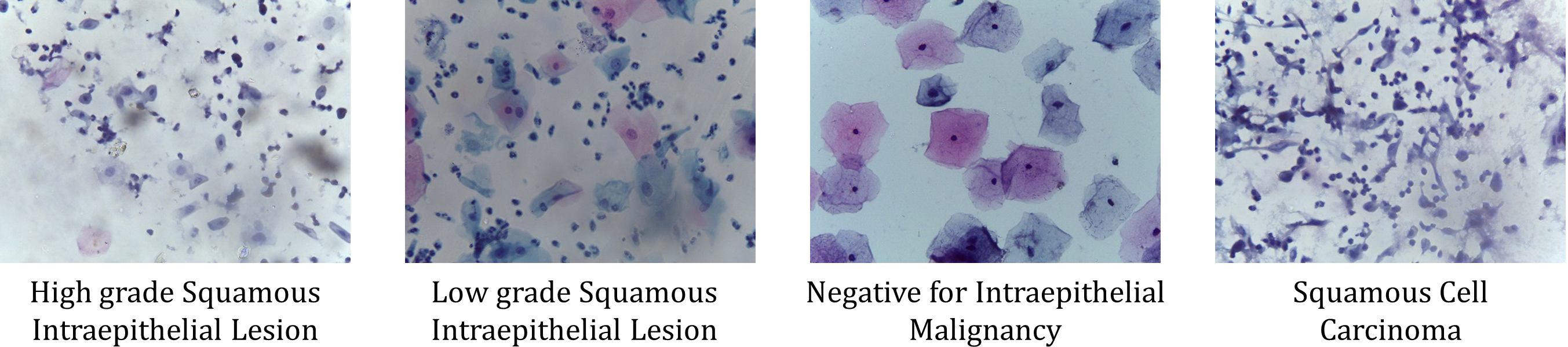}}\\
    \subfloat[SIPaKMeD Dataset]{\includegraphics[scale=0.5]{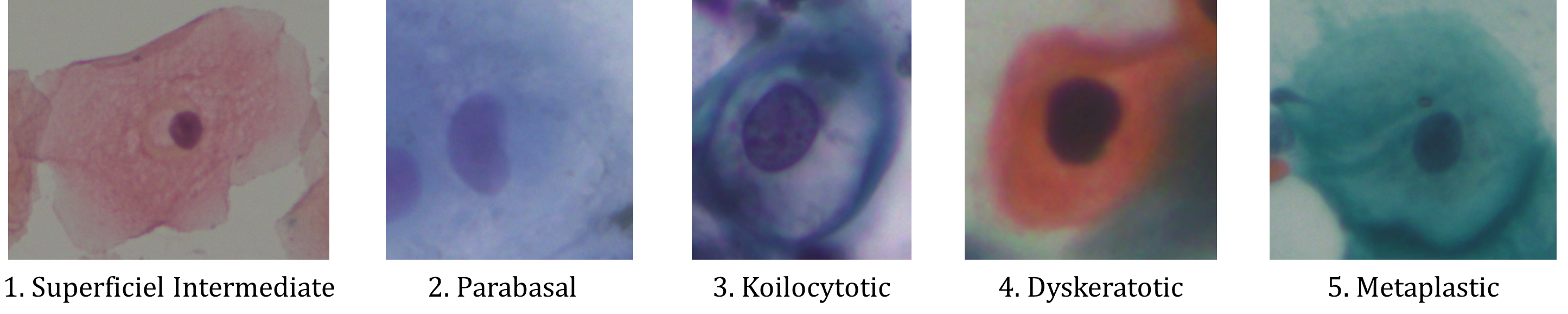}}
    \caption{Examples of images from each class in the three publicly available datasets}
    \label{data}
\end{adjustwidth}
\end{figure*}

We use three publicly available cervical cytology datasets in this study for evaluating the proposed classification framework:
\begin{enumerate}
    \item Herlev Pap Smear dataset by Jantzen et al. \cite{jantzen2005pap}
    \item Mendeley Liquid Based Cytology dataset by Hussain et al. \cite{hussain2020liquid}
    \item SIPaKMeD Pap Smear dataset by Plissiti et al. \cite{plissiti2018sipakmed}
\end{enumerate}
These datasets are described in brief in the following subsections.
\begin{table}[]
\centering
\caption{Distribution of images in the three publicly available datasets}
\label{data_dist}
\resizebox{\textwidth}{!}{
\begin{tabular}{|c|c|c|c|c|}
\hline
\textbf{Dataset} &
  \textbf{Class} &
  \textbf{Category} &
  \textbf{Cell type} &
  \textbf{Number of images} \\ \hline
\multirow{7}{*}{\textbf{\begin{tabular}[c]{@{}c@{}}Herlev\\ Pap Smear\\ (Total: 917)\end{tabular}}}    & 1 & Normal & Intermediate Squamous Epithelial        & 70  \\ \cline{2-5} 
 &
  2 &
  Normal &
  Columnar Epithelial &
  98 \\ \cline{2-5} 
 &
  3 &
  Normal &
  Superficial Squamous Epithelial &
  74 \\ \cline{2-5} 
 &
  4 &
  Abnormal &
  \begin{tabular}[c]{@{}c@{}}Mild Squamous \\ non-keratinizing Dysplasia\end{tabular} &
  182 \\ \cline{2-5} 
 &
  5 &
  Abnormal &
  \begin{tabular}[c]{@{}c@{}}Squamous cell carcinoma \\ in-situ intermediate\end{tabular} &
  150 \\ \cline{2-5} 
 &
  6 &
  Abnormal &
  \begin{tabular}[c]{@{}c@{}}Moderate Squamous \\ non-keratinizing Dysplasia\end{tabular} &
  146 \\ \cline{2-5} 
 &
  7 &
  Abnormal &
  \begin{tabular}[c]{@{}c@{}}Severe Squamous\\  non-keratinizing Dysplasia\end{tabular} &
  197 \\ \hline
\multirow{4}{*}{\textbf{\begin{tabular}[c]{@{}c@{}}Mendeley\\ LBC\\ (Total: 963)\end{tabular}}}        & 1 & Normal & Negative for Intraepithelial Malignancy & 613 \\ \cline{2-5} 
 &
  2 &
  Abnormal &
  \begin{tabular}[c]{@{}c@{}}Low grade Squamous\\ Intraepithelial Lesion (LSIL)\end{tabular} &
  163 \\ \cline{2-5} 
 &
  3 &
  Abnormal &
  \begin{tabular}[c]{@{}c@{}}High grade Squamous \\ Intraepithelial Lesion (HSIL)\end{tabular} &
  113 \\ \cline{2-5} 
 &
  4 &
  Abnormal &
  Squamous Cell Carcinoma (SCC) &
  74 \\ \hline
\multirow{5}{*}{\textbf{\begin{tabular}[c]{@{}c@{}}SIPaKMeD\\ Pap Smear\\ (Total: 4049)\end{tabular}}} & 1 & Normal & Superficial-Intermediate                & 831 \\ \cline{2-5} 
 &
  2 &
  Normal &
  Parabasal &
  787 \\ \cline{2-5} 
 &
  3 &
  Abnormal &
  Koilocytotic &
  825 \\ \cline{2-5} 
 &
  4 &
  Abnormal &
  Dyskeratotic &
  813 \\ \cline{2-5} 
 &
  5 &
  Benign &
  Metaplastic &
  793 \\ \hline
\end{tabular}
}
\end{table}
\subsubsection{Herlev Pap Smear Dataset}
The Herlev Pap Smear dataset is a publicly available benchmark dataset consisting of 917 single cell images distributed unevenly among 7 different classes. The distribution of images in each class are tabulated in Table \ref{data_dist}.
\subsubsection{Mendeley Liquid Based Cytology Dataset}
The Mendeley LBC dataset \cite{hussain2020liquid} developed at Obstetrics and Gynecology department of Guwahati Medical College and Hospital, consists of 963 whole slide images of cervical cytology distributed unevenly in four different classes as shown in Table \ref{data_dist}.
\subsubsection{SIPaKMeD Pap Smear dataset}
The SIPaKMeD Pap Smear dataset by Plissiti et al. \cite{plissiti2018sipakmed} consists of 4049 images of isolated cells (extracted from 966 whole slide images) categorized into five different classes based on their cytomorphological features. The distribution of images in the dataset is shown in Table \ref{data_dist}.
\subsection{Deep Features Extraction}
Handcrafted or manual feature extraction using traditional Machine Learning techniques has limitations both in terms of the number of features and their correlations. Extracting features from a large dataset is a tedious task and can incorporate human biases, affecting the quality of the features that can eventually affect the classification task. Redundant features might be extracted which might lead to higher rates of misclassification. So, in this work, we extract deep features from CNN classifiers. Deep Learning models use backpropagation to learn the important features themselves, and thus eliminates the tedious process of using handcrafted features. For the present study, we have used ResNet-50 \cite{he2016deep}, VGG-16 \cite{simonyan2014very}, Inception v3 \cite{szegedy2016rethinking} and DenseNet-121 \cite{huang2017densely} for extraction of features from the penultimate layer of the models.

While performing feature extraction from a CNN, we use the pre-trained model and fine-tune the CNN using our data, letting each image propagate through the layers in a forwarding direction, terminating at the pre-final layer, and taking out the output of this layer as the feature vector. We use pre-trained weights (Transfer Learning) in this study because the biomedical data is scarce and insufficient for Deep Learning models to work efficiently if trained from scratch. ImageNet \cite{deng2009imagenet} dataset consists of 14 billion images divided into 1000 classes. We use the models pretrained on this dataset and replace the final classification layer of size 1000 with a layer of size equals to the number of classes in our dataset. A model pretrained on such a large dataset already has learned important features from image data, and just needs fine-tuning for less number of epochs to train the final classification layer that we added.
\subsubsection{VGG-16}
The main characteristics of VGG nets \cite{simonyan2014very} includes the use of $3\times3$ convolution layers which gave a noticeable improvement in network performances while making the network deep. $3\times3$ receptive filters were used throughout the entire net with strides of $1$. Local Response Normalization (LRN) is not used in VGG Nets because memory consumption is more in such cases. The small-sized convolution filters give VGG Nets a chance to have a very large number of weight layers which in turn boosts performance. The input has a shape of $224\times224\times3$. In the present work, we fine-tune the VGG-16 model using our datasets, employing Stochastic Gradient Descent (SGD) optimizer and Rectified Linear Unit (ReLU) activation function.
\subsubsection{ResNet-50}
The ResNet-50 architecture \cite{he2016deep} consists of residual skip connections embedded that make the training of the network easier. The gradient vanishing problem is addressed at the same time due to the embedding of the skip connections, which allows very deep networks to be accommodated for a controlled computation cost. $224\times224\times3$ sized inputs are used in the ResNet-50 model, with SGD optimizer and ReLU activation function.
\subsubsection{Inception v3}
The salient feature of the Inception v3 architecture \cite{szegedy2016rethinking}, is the inception blocks that use parallel convolutions followed by channel concatenation. This leads to vivid features being extracted but with seemingly shallow networks. Parallel convolutions also allow the overfitting problem to be addressed while controlling the computational complexity. Inputs of shape $299\times299\times3$ are used with SGD optimizer and ReLU activation function for deep features extraction.
\subsubsection{DenseNet-121}
The DenseNet model \cite{huang2017densely} was proposed to address the vanishing gradient descent problem. The fundamental blocks in the DenseNet architecture are connected densely to each other leading to low computational requirement since the number of trainable parameters decreased heavily. The DenseNet architectures add small sets of feature maps owing to their narrow architecture. We used the DenseNet-121 variant using the SGD optimizer and ReLU activation function for deep features extraction.
\subsection{Principle Component Analysis}
Principal Component Analysis (PCA) is a linear dimensionality reduction method that transforms the higher dimensional data into a lower dimension by maximizing the variance of the lower dimension. PCA was first introduced by Wold et al. in 1987 [44], however, further development and implementation of PCA in machine learning problems was done quite significantly in the later period [45], [46]. The covariance matrix of the feature vector is computed first and followed by the computation of eigenvectors of this matrix. The eigenvectors that have the largest eigenvalues contribute to the formation of new reduced dimensionality of the feature vector. Thus, instead of losing some of the important features of the data, we kept the most important of the features by preserving 99\% of the variance.
Before applying the PCA algorithm for feature dimension reduction, we need to perform data preprocessing that is required for the further steps. Depending upon the $n$-dimensional training set $x^{(1)}, x^{(2)}, x^{(3)}, \hdots x^{(n)}$, we need to perform mean normalization or feature scaling similar to the supervised learning algorithms. The mean of each feature is computed as in Equation \ref{mean}
\begin{equation}\label{mean}
    \mu_i = \frac{1}{n}\sum_{j = 1}^{n}x_i^{(j)}
\end{equation}
Now, we replace each of the $x_i$ value with the $x_i - \mu_i$ value so that each of them has exactly zero mean value, however, if different features have different mean values, we can scale them so that they belong in a comparable range. In supervised learning, this scaling process of the $i^{th}$ element is defined by Equation \ref{scale}, where, $s_i$ is the $|max-mean|$ value or the static deviation of $i^{th}$ feature.
\begin{equation}\label{scale}
    x_i^{(j)}=\frac{x_i^{(j)}-\mu_i}{s_i}
\end{equation}
For reducing the dimension of the feature from $N$ to $m$ (where $m<N$), and to define the surface in $N$-dimensional space onto which we project the data, we need to find the mean square error of the projected data on the $m$ dimensional vector. The computational proof of the calculation of these m vectors: $u^{(1)}, u^{(2)},\hdots,u^{(m)}$ and the projected points: $zu^{(1)}, z^{(2)},\hdots,z^{(N)}$ on these vectors is complicated and beyond the scope of this paper. The covariance matrix is computed as in Equation \ref{covariance} where, the $x^{(j)}$ vector has $N\times1$ dimension and $(x^{(j)})^T$ has $1\times N$ dimension, thus making the covariance matrix of dimension of $N\times N$. Next, we calculate the eigenvalues and eigenvectors of the covariance matrix which represent the new magnitude of the feature vectors in the transformed vector space and their corresponding directions respectively.  The eigenvalues quantify the variance of all the vectors as we are dealing with the covariance matrix. If an eigenvector has high valued eigenvectors, that means that it has high variance and contains various important information about the dataset. On the other hand, eigenvectors will small eigenvalues contain very small information about the dataset.
\begin{equation}\label{covariance}
    covariance\;matrix = \frac{1}{N}\sum_{j=1}^{N}x^{(j)}\times (x^{(j)})^T
\end{equation}
Hence the $p^{th}$ complete principal component of a data vector $x^{(j)}$ in the transformed coordinates can be allocated a score $t^{(p)} = x^{(j)}\times w^{(p)}$ where $w^{(p)}$ is the $p^{th}$ eigenvector of the covariance matrix of $x^{(j)}$. Therefore the full PCA decomposition of the vector $X$ can be represented as $T=X\times W$, where $W$ is the eigenvector of the covariance matrix. Now, we need to select $m$-number of eigenvalues from these $N$ eigenvectors by maximizing the variance of the preserved original data while reducing the total square reconstruction error. Next, we calculate the Cumulative Explained Variance (CEV) which is the sum of variances (information) containing in the top $m$ principal components. Then we set a threshold value above which, the eigenvalues will be considered as useful and the rest will be discarded as unimportant features. For our experiment we have set the threshold value to 99, meaning that we have kept 99\% of the variance of the data retained in the reduced feature vector. As different CNN extracts features of different modalities, the number of selected features after PCA and GWO are different based on the feature distribution in those feature sets. 

The pseudo-code of dimensionality reduction using PCA is shown in Algorithm \ref{algopca}.

\begin{algorithm}
    {\small
    \textbf{define\_function: PCA}\\
    {\em Input:}\\
    Feature set $X$ of dimension $d$
    \begin{algorithmic}
        \STATE Compute Co-variance matrix $\Pi$
        \STATE\WHILE{(i$\leq$ d)}
            \STATE\WHILE{(j $\leq$ d)}
               \STATE $\mu_i \longleftarrow$ sample mean of feature $i$ 
               \STATE $\mu_j \longleftarrow$ sample mean of feature $j$
               \STATE $\sigma_{ij}=\frac{1}{n}\sum\limits_{k=1}^n(x_i^k-\mu_i)(x_j^k-\mu_j)$
               \STATE j=j+1
            \ENDWHILE
            \STATE $i=i+1$
        \ENDWHILE
        \STATE decompose $\pi$ into eigenvalues and eigenvectors
        \STATE calculate cumulative explained variance (CEV)
        \STATE\IF{(CEV$\geq$ threshold)}
            \STATE Construct projection matrix W
        \ENDIF
        \STATE transform input $X$ using $W$
        \STATE obtain k-dimensional feature subspace $X'$
        \STATE {\em return} $X'$
        
    \end{algorithmic}
    }
\caption{Pseudo-code for Principal Component Analysis}
\label{algopca}
\end{algorithm}
\subsection{Grey Wolf Optimizer}
Grey Wolf Optimization (GWO) \cite{mirjalili2014grey} is a nature-inspired meta-heuristic optimization algorithm that mimics the leadership hierarchy of Grey Wolf ($Canis$ $lupus$) and their hunting process for the optimization. Four types of GWO agents are deployed for simulation of the optimization algorithm named alpha, beta, delta, and omega. They mimic the three-step hunting methods of the grey wolf: finding the prey, encircling them, and finally attacking them for the sake of optimization. 

The grey wolves follow the leadership of the alpha wolf, which is the topmost category of their strict social hierarchy. The alpha wolf is not necessarily the strongest and the fittest ones, but they can maintain the discipline of the whole pack. The major decisions are taken by the alpha wolves but often accompanied by the subordinates, the beta wolf. They are the next lower level of wolves in their social hierarchy and convey the decisions o the alpha to the lower levels of wolves. They are generally the fittest candidates for alpha if the alpha becomes old or weak and plays a major role in maintaining the pack as a whole. The next level of wolves is called delta and they play a very major role in decision-making and other important activities of the pack. The last and the least important category of the pack is named omega and they often play the role of scapegoat in society. Thus the complete pack is formed based on dominance hierarchy. The mathematical model for the steps of optimization that mimics their hunting process is described below.
\subsubsection{Social Hierarchy}
Similar to the social hierarchy of grey wolf, the optimizer allocates the three fittest solutions as alpha, beta, and delta and the rest of the search agents are bound to arrange them and adjust accordingly as the parameters of the alpha, beta, and delta wolves. These three wolves are followed by the omega wolves.
\subsubsection{Encircling the Prey (Optimal Solution)}
To mathematically represent the encircling of prey, Equations \ref{enc1}, and \ref{enc2} are used where $t$ is the present iteration and $\Vec{C}$ and $\Vec{A}$ are the coefficient vectors, $\Vec{x_p}$ indicates the vector position of the prey and $\Vec{x}$ indicates the vector position of the grey wolf.
\begin{equation}\label{enc1}
    \Vec{D}=|\Vec{C}\times\Vec{x_p}(t)-\Vec{x}(t)|
\end{equation}
\begin{equation}\label{enc2}
    \Vec{x}(t+1)=\Vec{x_p}(t)-\Vec{A}\times\Vec{D}
\end{equation}
The expression for $\Vec{A}$ and $\Vec{C}$ areas in Equations \ref{vec_a} and \ref{vec_c} respectively, where $\Vec{r_1}$ and $\Vec{r_2}$ are the random valued vectors between $0$ and $1$ inclusive and the value of $\Vec{a}$ decreases from $2$ to $0$ linearly with increase in iteration. The two random variables $\Vec{r_1}$ and $\Vec{r_2}$ allow a grey wolf agent to reach any position between the points. The agents of the grey wolf algorithm can position themselves around the fittest solution by adjustment of the value of $\Vec{C}$ and $\Vec{A}$. 
\begin{equation}\label{vec_a}
    \Vec{A}=2\times\Vec{a}\times\Vec{r_1}-\Vec{a}
\end{equation}
\begin{equation}\label{vec_c}
    \Vec{C}=2\times\Vec{r_2}
\end{equation}
The same is applicable for $n$-dimensional optimization where the grey wolf agents move along the hyper-sphere or hyper-cubes around the fittest solution obtained.
\subsubsection{Reaching the Optimal Solution}
In real life, grey wolves hunt for their prey being led by the alpha who is accompanied by beta and delta wolves and the other wolves follow their instructions. To simulate this hunting principle, we allow some random valued agents and find their fitness and consider the three most accurate results as alpha, beta, and delta as in abstract search space we have no idea about the position of the agents and the prey.  The rest of the agents including the omega wolves are bound to change their positions and orientations according to the three best wolf agents.
\begin{equation}
    D_{\alpha} = |\Vec{C_1}\times\Vec{x_{\alpha}}-\Vec{x}|
\end{equation}
\begin{equation}
    D_{\beta} = |\Vec{C_2}\times\Vec{x_{\beta}}-\Vec{x}|
\end{equation}
\begin{equation}
    D_{\delta} = |\Vec{C_3}\times\Vec{x_{\delta}}-\Vec{x}|
\end{equation}
\begin{equation}\label{update1}
    \Vec{x_{1}} = \Vec{x_{\alpha}}-\Vec{A_1}\times \Vec{D_{\alpha}}
\end{equation}
\begin{equation}\label{update2}
    \Vec{x_{2}} = \Vec{x_{\beta}}-\Vec{A_2}\times \Vec{D_{\beta}}
\end{equation}
\begin{equation}\label{update3}
    \Vec{x_{3}} = \Vec{x_{\delta}}-\Vec{A_3}\times \Vec{D_{\delta}}
\end{equation}
\begin{equation}
    \Vec{x}(t+1) = \frac{\Vec{x_1}+\Vec{x_2}+\Vec{x_3}}{3}
\end{equation}
The search agents update their position by these equations, however, the final position of the agents are not predefined, rather they are the random positions according to the position of the alpha, beta, and delta agents and within a certain circle which is determined by the position of the three best-fit solutions.
\subsubsection{Exploiting the Prey}
Grey wolves encircle their prey until the prey stops movement and this freezing the prey is known as exploiting. In the mathematical model, the value of $\Vec{a}$ is decreased with the agents approaching the prey, and hence the value of $\Vec{A}$ is modified further. The fluctuation of $\Vec{A}$ is stopped as the value of $\Vec{A}$ changes from $\Vec{-2a}$ to $\Vec{+2a}$. The value of $\Vec{a}$ changes from $2$ to $0$ with the increase in iterations. The search agents can take any position between their current position and the position of the prey as the alpha, beta, and delta wolves approach the prey for hunting.
\subsubsection{Exploring for the prey}
The grey wolf agents diverge in search of prey and they finally converge for attacking the prey. In mathematical modelling, this phenomenon is regulated by the value of $\Vec{A}$; if the value of $\Vec{A}$ is greater than $1$ or less than $-1$, the grey wolf agents diverge from each other and find for some more suitable prey. However, if $\Vec{A}$ has a value between $-1$ and $+1$, the agents converge towards the prey.

\begin{figure}
    \centering
    \includegraphics[scale=0.7]{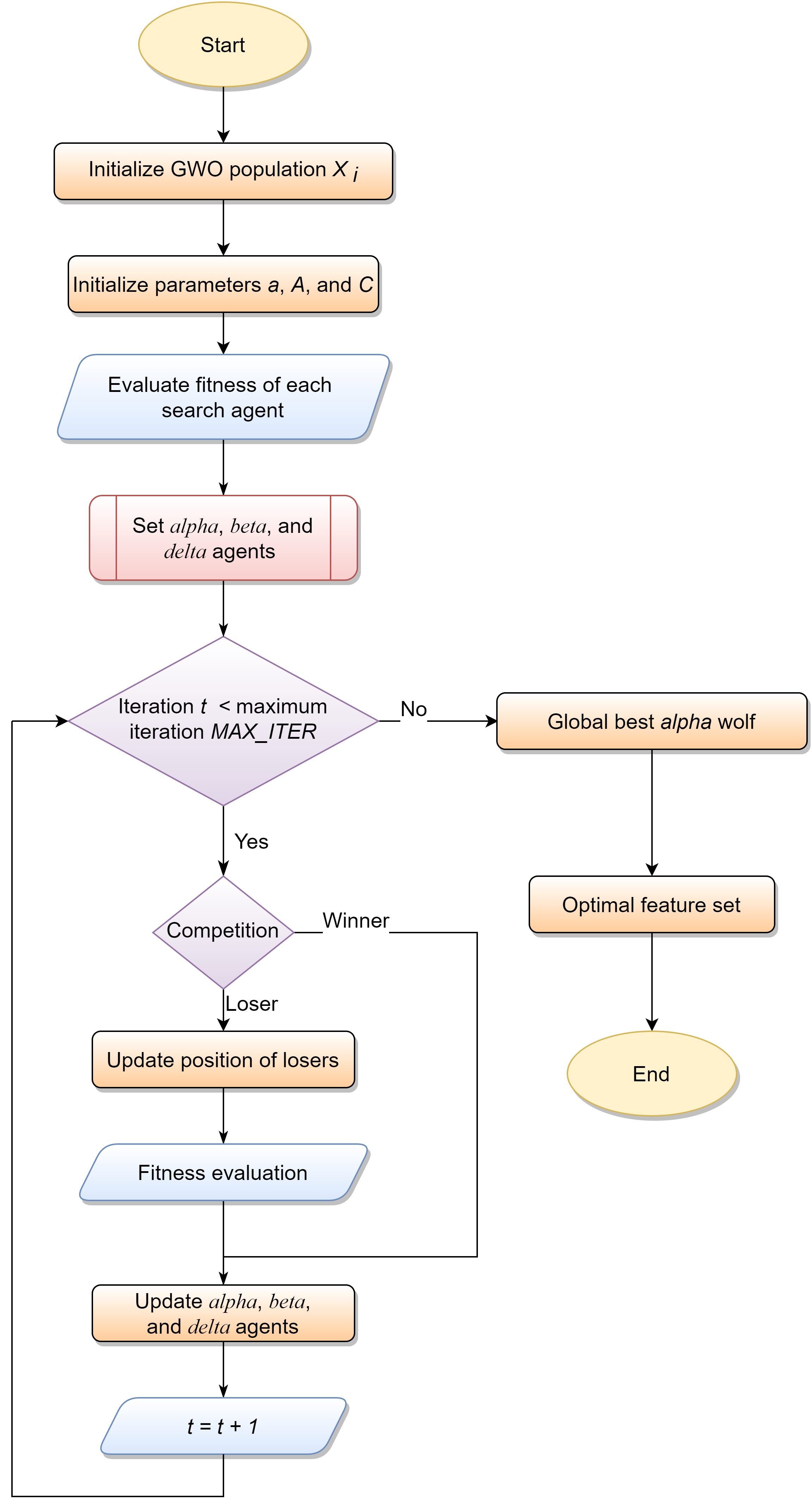}
    \caption{\textcolor{black}{Flowchart showing the workflow of the Grey Wolf Optimization algorithm used in the proposed framework.}}
    \label{flowchart}
\end{figure}

The overall pseudo-code of the GWO algorithm is shown in Algorithm \ref{algogwo} and the flowchart for the algorithm is shown in Figure \ref{flowchart}.
\begin{algorithm}[h]
    {\small
    \textbf{define\_function: GWO}\\
    {\em Input:}\\
    Number of Search Agents: $n$\\
    Maximum number of iterations: $MAX\_ITER$
    \begin{algorithmic}[]
    \STATE Initialize the GWO population $X_i\:\: \forall i=1,2,3,....,n$
    \STATE Initialize a, A, and C \tcp{According to Equations \ref{vec_a},\ref{vec_c}}
    \STATE Calculate the fitness of each search agent
    \STATE $X_{\alpha}$ = Alpha wolf (Best search agent)
    \STATE $X_{\beta}$ = Beta wolf (Second best search agent)
    \STATE $X_{\delta}$ = Delta wolf (Third best search agent)
        \STATE\WHILE{t $<$ $MAX\_ITER$}
            \STATE\FOR{each search agent}
                \STATE Update position of current search agent \tcp{According to Equation \ref{enc2}}
            \ENDFOR
            \STATE Update a, A, and C
            \STATE Calculate the fitness of each search agent
            \STATE Update $X_{\alpha}, X_{\beta}, X_{\delta}$ \tcp{According to Equations \ref{update1}, \ref{update2} \& \ref{update3}}
            \STATE $t=t+1$
        \ENDWHILE
    \STATE {\em return} $X_{\alpha}$
    \end{algorithmic}
    
    \caption{Pseudo-code for the Grey Wolf Optimizer for feature selection.}
    \label{algogwo}
    }
\end{algorithm}

\subsection{Classification}
After optimization, the final step is to fit the selected features to the classifier for the classification task. Due to a large number of features in some cases, we used incremental learning where a small batch of the dataset is selected for training the classifier and the loop over all the dataset and continue training until we reach convergence. This is fast and computationally efficient. We used an SVM classifier with the ‘RBF’ kernel for the multi-class classification task.
\begin{algorithm}
    {\em Input:}\\
    Raw RGB images: $I_{RGB}$
    \begin{algorithmic}
        \STATE $f_i$=deep features extracted from $i^{th}$ CNN; $i={1,2,3,4}$
        \STATE $f_{c}=concat(f_1,f_2,\hdots,f_i)$;$i={1,2,3,4}$  \tcp{concatenation of features}
        \STATE $F_{PCA}=PCA(f_{c})$ \tcp{Using algorithm \ref{algopca}}
        \STATE $F_{GWO}=GWO(F_{PCA})$ \tcp{Using Algorithm \ref{algogwo}}
        \STATE train-test split$\to train_{F_{GWO}},test_{F_{GWO}}$
        \STATE $clf=classifier.fit(train_{F_{GWO}})$ \tcp{Train SVM Classifier}
        \STATE $predicted=clf.predict(test_{F_{GWO}})$ \tcp{Make predictions on test set}
        \STATE Compare predictions and labels and evaluate performance
    \end{algorithmic}
    \caption{Pseudo-code for overall workflow.}
    \label{algooverall}
\end{algorithm}
\subsubsection{Support Vector Machine}
SVM \cite{zhang2012support} is a supervised learning model, which, in a set of training examples, properly labelled with different classes, add new examples to each class making a complete non-probabilistic binary classifier out of this SVM, and is associated with some typical learning algorithms which analyse the data, specifically used for regression and classification tasks. SVM model representation of the training samples in the feature plane is such that a separation between the examples belonging to different classes becomes so prominent, that a curve can be fit in that space between two classes which maintain maximum distances from every point of each class and SVM fits that curve. %
\section{Results and Discussion}\label{results}
\begin{table*}[]
\centering
\caption{Reduction in feature dimension and improvements in training time after principal component analysis on the Herlev dataset}
\label{reduction_features}
\resizebox{\textwidth}{!}{
\begin{tabular}{|c|c|c|c|c|}
\hline
\textbf{\begin{tabular}[c]{@{}c@{}}Model used for\\ feature extraction\end{tabular}} &
  \textbf{\begin{tabular}[c]{@{}c@{}}No. of features \\ (before PCA)\end{tabular}} &
  \textbf{\begin{tabular}[c]{@{}c@{}}No. of features\\ (after PCA)\end{tabular}} &
  \textbf{\begin{tabular}[c]{@{}c@{}}Reduction in \\ feature dimension (\%)\end{tabular}} &
  \textbf{\begin{tabular}[c]{@{}c@{}}Improvement in\\ average training time (\%)\end{tabular}} \\ \hline
ResNet-50 \cite{he2016deep}                                                                              & 100353 & 383 & 99.62 & 88.425 \\ \hline
VGG-16 \cite{simonyan2014very}                                                                                & 25088  & 364 & 98.55 & 85.215 \\ \hline
DenseNet-121 \cite{huang2017densely}                                                                           & 50177  & 330 & 99.34 & 81.449 \\ \hline
Inception v3 \cite{szegedy2016rethinking}                                                                      & 131073 & 325 & 99.75 & 84.228 \\ \hline
ResNet-50+VGG-16                                                                        & 125441 & 456 & 99.63 & 80.221 \\ \hline
DenseNet-121+Inception v3                                                           & 181250 & 687 & 99.62 & 82.694 \\ \hline
\begin{tabular}[c]{@{}c@{}}ResNet-50+VGG-16+\\ DenseNet-121+Inception v3\end{tabular} & 306691 & 796 & 99.74 & 85.737 \\ \hline
\end{tabular}
}
\end{table*}

\begin{figure*}
    \centering
    \subfloat[Herlev Pap Smear dataset]{\includegraphics[scale=0.45]{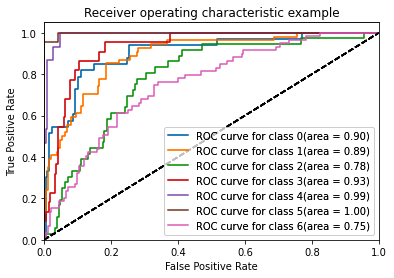}}
    \subfloat[Mendeley LBC dataset]{\includegraphics[scale=0.45]{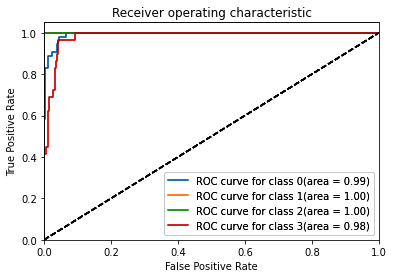}}\\
    \subfloat[SIPaKMeD Pap Smear dataset]{\includegraphics[scale=0.45]{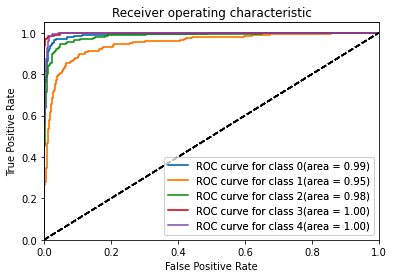}}
    \caption{ROC curves obtained by the proposed method for the three datasets: (a) Herlev Pap Smear dataset (b) Mendeley LBC dataset and (c) SIPaKMeD Pap Smear dataset.}
    \label{roc}
\end{figure*}

After extracting the features from the dataset using the CNN architectures said in Section \ref{methods}, and the features were concatenated. We then use PCA (which retained 99\% of the variance of the data) for the reduction in the dimensionality of the feature space and improvements in feature qualities respectively. Table \ref{reduction_features} shows the statistics of reduction in feature dimensionality as well as the improvement of training time after this for the Herlev dataset. Then, we used the GWO algorithm and finally split the dataset and calculated the accuracy score for the training, validation, and testing sets. The overall workflow is shown in the form of pseudo-code in Algorithm \ref{algooverall}. The results of our experiments are discussed in this section.

The metrics used for performance evaluation of the classification task for the multi-class problem is calculated based on Equations \ref{acc}, \ref{pre}, \ref{rec}, \ref{f1} which are derived from a confusion matrix $M$.
\begin{equation}
Accuracy = \frac{\sum_{i=1}^{N}M\textsubscript{ii}}{\sum_{i=1}^{N}\sum_{j=1}^{N}M\textsubscript{ij}}
\label{acc}
\end{equation}
\begin{equation}
Precision\textsubscript{i} = \frac{M\textsubscript{ii}}{\sum_{j=1}^{N}M\textsubscript{ji}}
\label{pre}
\end{equation}
\begin{equation}
Recall\textsubscript{i} = \frac{M\textsubscript{ii}}{\sum_{j=1}^{N}M\textsubscript{ij}}
\label{rec}
\end{equation}
\begin{equation}
F1-Score\textsubscript{i} = \frac{2}{\frac{1}{Precision\textsubscript{i}}+\frac{1}{Recall\textsubscript{i}}}
\label{f1}
\end{equation}
To cross-validate the results of the classification task on different datasets and different features, we performed an AUC-ROC test on different datasets. The ROC (Receiver Operating Characteristics) curve is an important analyzing tool for validating the clinical findings of our experiment. The different line segments in the OVA (One Vs. All) ROC represent different classes stating that how good the features and the classifier performance are for classifying the different classes which can be broadly categorized in normal and infected cases. It represents the graphical analysis of the TPR (True Positive Rate) against the FPR (False Positive Rate) as the two operating characteristics criterion of the classifier based on the features selected. A false-positive result is a case when data of a healthy or uninfected class is predicted as an unhealthy or infected case by a classifier and it’s a major drawback of the classification task. This is reciprocated by the points lying far above the diagonal line of the ROC curve suggesting that the TPR is significantly high as compared to FPR. Another important feature for analyzing the classification result is the AUC (Area Under Curve) of the ROC curve which was computed considering the 97\% of the confidence interval. The analysis using the AUC-ROC curves for different datasets and different features are discussed further.

\subsection{Results on Herlev Pap Smear Dataset}
\begin{figure*}
\begin{adjustwidth}{-2cm}{}
    \centering
    \includegraphics[scale=0.5]{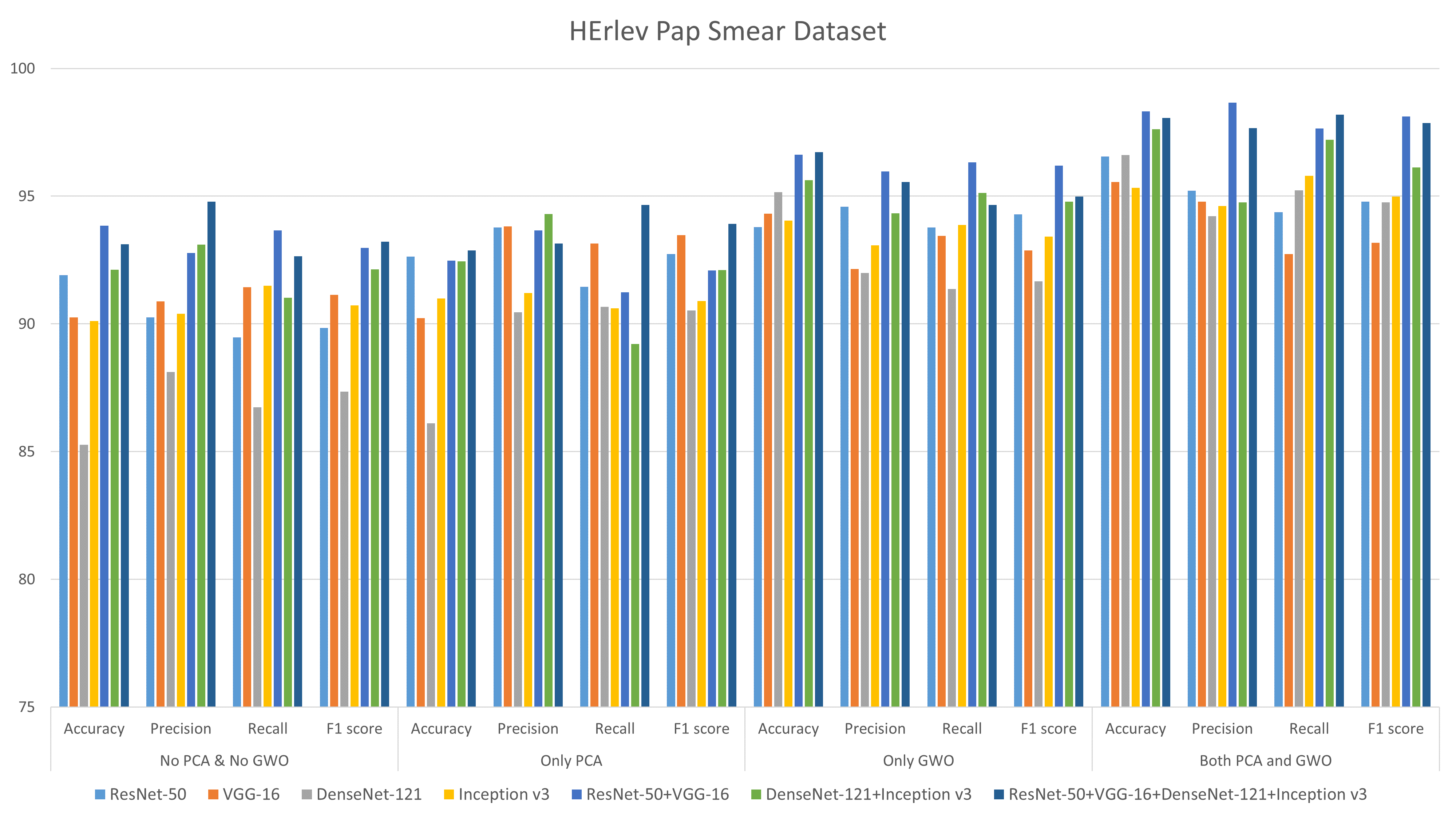}
    \caption{Results on the Herlev Pap Smear dataset}
    \label{res_Herlev}
\end{adjustwidth}
\end{figure*}

The results obtained on different experiments on the Herlev Pap Smear dataset are shown in Figure \ref{res_Herlev}. The best classification results observed this dataset was achieved by merging the feature extracted from ResNet-50 and VGG-16 models, which gave the performance metrics as follows: Accuracy = 98.32\%, Precision = 98.66\%, Recall = 97.65\% and F1-score = 98.12\%.

\subsection{Results on the Mendeley LBC Dataset}
\begin{figure*}
\begin{adjustwidth}{-2cm}{}
    \centering
    \includegraphics[scale=0.5]{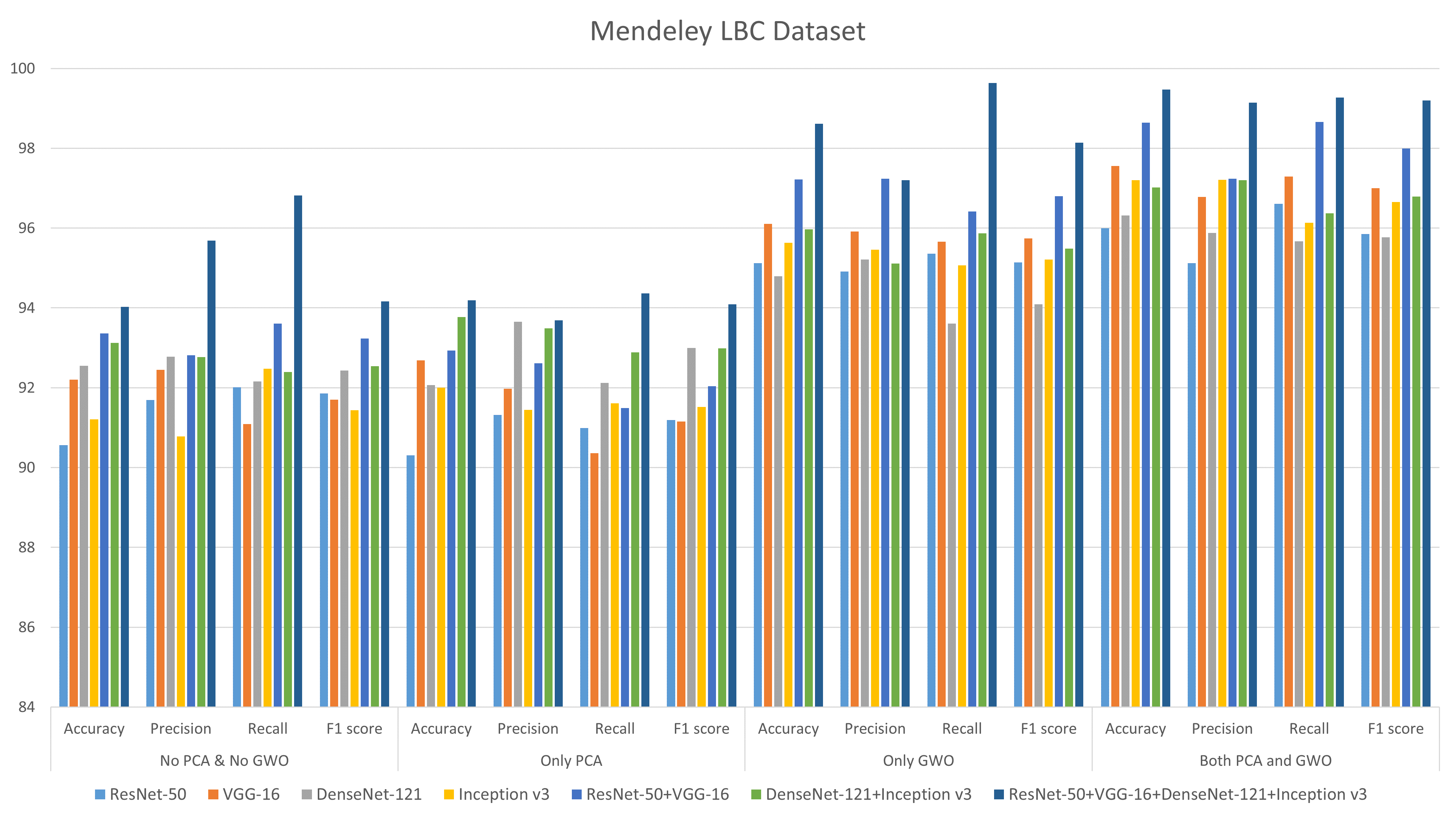}
    \caption{Results on the Mendeley LBC dataset}
    \label{res_mendeley}
\end{adjustwidth}
\end{figure*}

The results obtained on different experiments on the Mendeley LBC dataset are shown in Figure \ref{res_mendeley}. The best results on this dataset are obtained by merging features extracted from VGG-16, ResNet-50, Inception v3 and DenseNet-121: Accuracy = 99.47\%, Precision = 99.14\%, Recall = 99.27\% and F1-score = 99.20\%.

\subsection{Results on SIPaKMeD Pap Smear dataset}
\begin{figure*}
\begin{adjustwidth}{-2cm}{}
    \centering
    \includegraphics[scale=0.5]{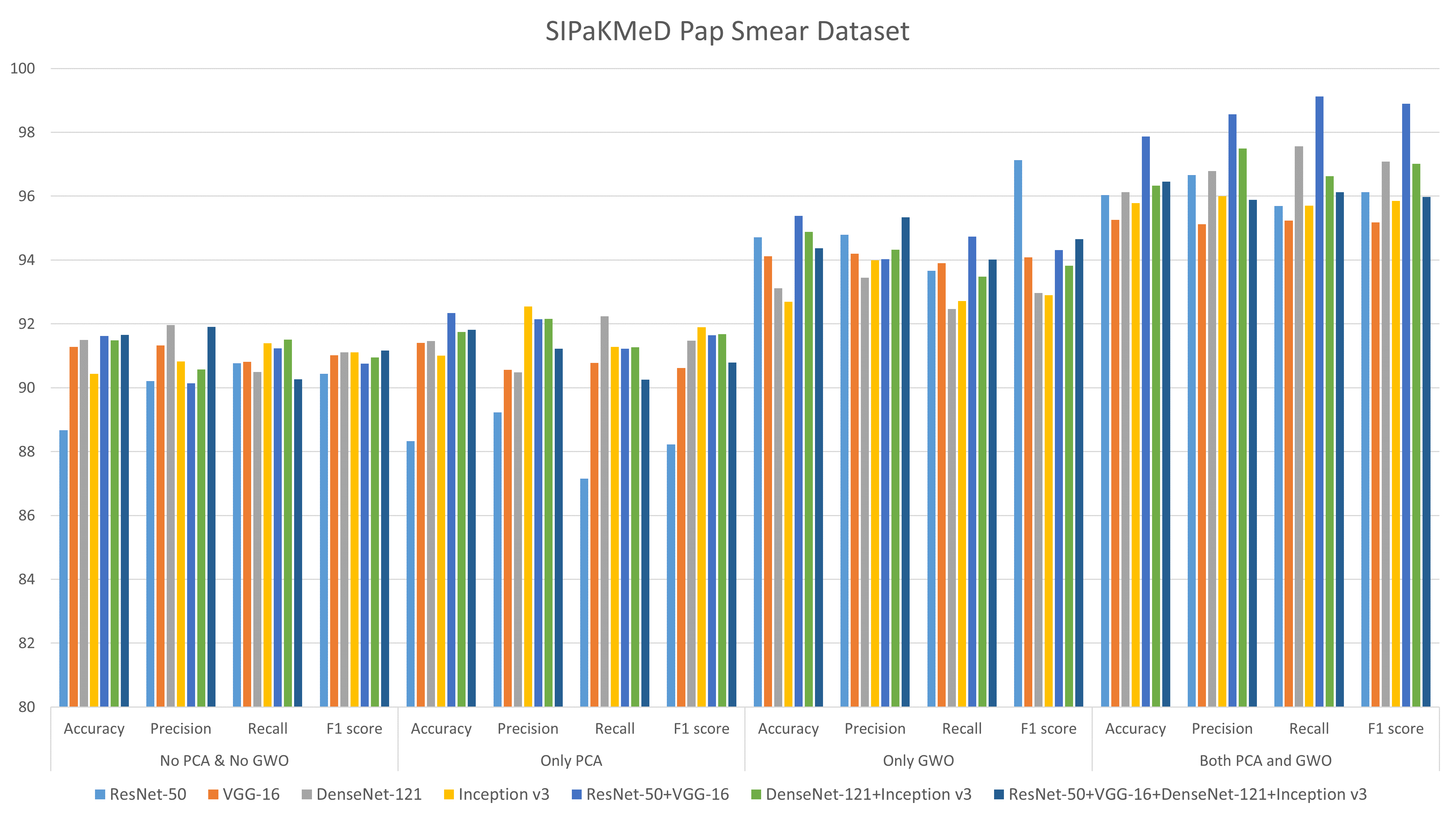}
    \caption{Results on the SIPaKMeD Pap Smear Dataset}
    \label{res_sipak}
\end{adjustwidth}
\end{figure*}

The results obtained on different experiments on the SIPaKMeD Pap Smear dataset are shown in Figure \ref{res_sipak}. The best results on the dataset are obtained by merging features extracted from VGG-16 and ResNet-50: Accuracy = 97.87\%, Precision = 98.56\%, Recall = 99.12\% and F1-score = 98.89\%.

\begin{table*}[]
\centering
\caption{Accuracies and Losses on training, validation and testing sets using both PCA and GWO on the three datasets dataset (all the accuracy measurements are in \% and measured after 30 epochs)}
\label{acc_loss}
\resizebox{\textwidth}{!}{
\begin{tabular}{|c|c|c|c|c|c|c|c|}
\hline
\textbf{Dataset} &
  \textbf{Feature extractor model} &
  \textbf{\begin{tabular}[c]{@{}c@{}}Training \\      Accuracy\end{tabular}} &
  \textbf{\begin{tabular}[c]{@{}c@{}}Training \\      Loss\end{tabular}} &
  \textbf{\begin{tabular}[c]{@{}c@{}}Validation \\      Accuracy\end{tabular}} &
  \textbf{\begin{tabular}[c]{@{}c@{}}Validation \\      Loss\end{tabular}} &
  \textbf{\begin{tabular}[c]{@{}c@{}}Testing \\      Accuracy\end{tabular}} &
  \textbf{\begin{tabular}[c]{@{}c@{}}Testing \\      Loss\end{tabular}} \\ \hline
\multirow{7}{*}{\textbf{\begin{tabular}[c]{@{}c@{}}Herlev\\ Pap Smear\end{tabular}}} &
  ResNet-50 \cite{he2016deep}&
  97.77 &
  0.026 &
  96.33 &
  0.032 &
  96.55 &
  0.028 \\ \cline{2-8} 
 &
  VGG-16 \cite{simonyan2014very}&
  96.36 &
  0.031 &
  94.21 &
  0.109 &
  95.56 &
  0.081 \\ \cline{2-8} 
 &
  DenseNet-121 \cite{huang2017densely}&
  97.89 &
  0.02 &
  97.01 &
  0.024 &
  96.61 &
  0.024 \\ \cline{2-8} 
 &
  Inception v3 \cite{szegedy2016rethinking}&
  96.33 &
  0.032 &
  95.17 &
  0.098 &
  95.32 &
  0.094 \\ \cline{2-8} 
 &
  \textbf{ResNet-50+VGG-16} &
  \textbf{98.77} &
  \textbf{0.011} &
  \textbf{98.00} &
  \textbf{0.019} &
  \textbf{98.32} &
  \textbf{0.016} \\ \cline{2-8} 
 &
  DenseNet-121+Inception v3 &
  97.91 &
  0.026 &
  96.01 &
  0.031 &
  97.62 &
  0.027 \\ \cline{2-8} 
 &
  \begin{tabular}[c]{@{}c@{}}ResNet-50+VGG-16+\\ DenseNet-121+Inception v3\end{tabular} &
  98.3 &
  0.018 &
  97.95 &
  0.021 &
  98.06 &
  0.019 \\ \hline
\multirow{7}{*}{\textbf{\begin{tabular}[c]{@{}c@{}}Mendeley\\ LBC\end{tabular}}} &
  ResNet-50 \cite{he2016deep}&
  96.88 &
  0.066 &
  96.11 &
  0.071 &
  96 &
  0.079 \\ \cline{2-8} 
 &
  VGG-16 \cite{simonyan2014very}&
  97.91 &
  0.051 &
  96.39 &
  0.068 &
  97.56 &
  0.059 \\ \cline{2-8} 
 &
  DenseNet-121 \cite{huang2017densely}&
  97.16 &
  0.061 &
  96.15 &
  0.07 &
  96.32 &
  0.071 \\ \cline{2-8} 
 &
  Inception v3 \cite{szegedy2016rethinking}&
  97.5 &
  0.058 &
  97.05 &
  0.06 &
  97.2 &
  0.061 \\ \cline{2-8} 
 &
  ResNet-50+VGG-16 &
  99.04 &
  0.039 &
  97.96 &
  0.054 &
  98.64 &
  0.054 \\ \cline{2-8} 
 &
  DenseNet-121+Inception v3 &
  98.06 &
  0.044 &
  96.49 &
  0.067 &
  97.02 &
  0.064 \\ \cline{2-8} 
 &
  \textbf{\begin{tabular}[c]{@{}c@{}}ResNet-50+VGG-16+\\ DenseNet-121+Inception v3\end{tabular}} &
  \textbf{99.58} &
  \textbf{0.03} &
  \textbf{98.88} &
  \textbf{0.043} &
  \textbf{99.47} &
  \textbf{0.04} \\ \hline
\multirow{7}{*}{\textbf{\begin{tabular}[c]{@{}c@{}}SIPaKMeD\\ Pap Smear\end{tabular}}} &
  ResNet-50 \cite{he2016deep}&
  96.85 &
  0.028 &
  96.77 &
  0.049 &
  96.03 &
  0.048 \\ \cline{2-8} 
 &
  VGG-16 \cite{simonyan2014very}&
  96.71 &
  0.03 &
  94.02 &
  0.071 &
  95.26 &
  0.059 \\ \cline{2-8} 
 &
  DenseNet-121 \cite{huang2017densely}&
  96.31 &
  0.035 &
  96.04 &
  0.058 &
  96.12 &
  0.046 \\ \cline{2-8} 
 &
  Inception v3 \cite{szegedy2016rethinking}&
  96.02 &
  0.039 &
  95.91 &
  0.06 &
  95.78 &
  0.055 \\ \cline{2-8} 
 &
  \textbf{ResNet-50+VGG-16} &
  \textbf{98.48} &
  \textbf{0.014} &
  \textbf{97.55} &
  \textbf{0.041} &
  \textbf{97.87} &
  \textbf{0.034} \\ \cline{2-8} 
 &
  DenseNet-121+Inception v3 &
  97.32 &
  0.02 &
  95.39 &
  0.066 &
  96.33 &
  0.044 \\ \cline{2-8} 
 &
  \begin{tabular}[c]{@{}c@{}}ResNet-50+VGG-16+\\ DenseNet-121+Inception v3\end{tabular} &
  96.92 &
  0.025 &
  96.66 &
  0.051 &
  96.46 &
  0.042 \\ \hline
\end{tabular}
}
\end{table*}

\subsection{Comparison with Existing Literature}
\begin{table*}[tbp]
\centering
\caption{Comparison (ACC, in \%) with standard optimization algorithms: PSO = Particle Swarm Optimization \cite{kennedy1995particle}; MVO = Mean Variance Optimization \cite{erlich2010mean}; GWO = Grey Wolf Optimizer \cite{mirjalili2014grey}; MFO = Moth Flame Optimization \cite{mirjalili2015moth}; WOA = Whale Optimization Algorithm \cite{mirjalili2016whale}; FFA = Firefly Algorithm \cite{yang2009firefly}; BAT = Bat Optimization Algorithm \cite{yang2012bat}; GA = Genetic Algorithm \cite{de1975analysis, holland1992adaptation}}
\label{tableoas}
\resizebox{\textwidth}{!}{
\begin{tabular}{|l|c|c|c|c|c|c|}
\hline
Optimization & \multicolumn{2}{c|}{\textbf{Mendeley LBC Dataset}}                   & \multicolumn{2}{c|}{\textbf{Herlev Pap-smear Dataset}}           & \multicolumn{2}{c|}{\textbf{SIPaKMeD 5-class Dataset}}           \\ \cline{2-7} 
Alogotithms                                                  & \textbf{ACC} & \textbf{\# of Features} & \textbf{ACC} & \textbf{\# of Features} & \textbf{ACC} & \textbf{\# of Features} \\ \hline \hline
PSO                                      & 95.90            & 920                                  & 92.58            & 992                                 & 90.14            & 1014                                 \\ 
MVO                                      & 96.91            & 720                                  & 94.26            & 764                                  & 90.48            & 843                                  \\ 
GWO                                      & 92.14            & 810                                  & 92.40            & 807                                  & 89.98            & 791                                  \\ 
MFO                                      & 94.20             & 803                                  & 93.19             & 851                                  & 90.58            & 832                                  \\ 
WOA                                      & 95.08            & 843                                  & 92.36            & 847                                  & 90.58            & 802                                  \\ 
FFA                                      & 94.42            & 715                                  & 92.46            & 820                                  & 89.56            & 792                                  \\ 
BAT                                      & 95.64             & 857                                 & 94.58            & 762                                  & 90.21            & 749                                  \\ 
GA                                      & 98.23             & 724                                  & 95.26            & 784                                  & 95.43            & 796                                  \\ 
\textbf{PCA+GWO}                         & \textbf{99.47}    & \textbf{762}                         & \textbf{98.32}    & \textbf{796}                       & \textbf{97.87}    & \textbf{736}                         \\ \hline
\end{tabular}
}
\mbox{}\\
\end{table*}

Several models have been proposed in the literature for cervical cell classification as discussed in Section~\ref{relwork}. Our proposed work and the results achieved are therefore compared with some of these models that used the same datasets to assess the reliability of the proposed framework and the results are tabulated in Table \ref{comparison}. No papers as of yet have been published that use the Mendeley LBC dataset, and thus we are unable to compare our method in that dataset.
\begin{table}[]
\centering
\caption{Comparison of the proposed method with existing literature}
\label{comparison}
\resizebox{\textwidth}{!}{
\begin{tabular}{|c|c|c|}
\hline
\textbf{Dataset} &
  \textbf{Method} &
  \textbf{Results} \\ \hline
\multirow{5}{*}{\textbf{\begin{tabular}[c]{@{}c@{}}Herlev \\ Pap Smear\end{tabular}}} &
  Genctav et al. \cite{gencctav2012unsupervised} &
  \begin{tabular}[c]{@{}c@{}}Precision: 88\%±0.15\\      Recall: 93\%±0.15\end{tabular} \\ \cline{2-3} 
 &
  Bora et al. \cite{bora2017automated} &
  Accuracy: 96.51\% \\ \cline{2-3} 
 &
  Win et al. \cite{win2020computer} &
  Accuracy: 90.84\% \\ \cline{2-3} 
 &
  Chankong et al. \cite{chankong2014automatic} &
  Accuracy: 93.78\% \\ \cline{2-3} 
 &
  \textbf{Proposed Method} &
  \textbf{\begin{tabular}[c]{@{}c@{}}Accuracy: 98.32\%\\      Precision: 98.66\%\\      Recall: 97.65\%\\      F1-score: 98.12\%\end{tabular}} \\ \hline
\multirow{3}{*}{\textbf{\begin{tabular}[c]{@{}c@{}}SIPaKMeD \\ Pap Smear\end{tabular}}} &
  Win   et al. \cite{win2020computer} &
  Accuracy: 94.09\% \\ \cline{2-3} 
 &
  Plissiti et al. \cite{plissiti2018sipakmed} &
  \begin{tabular}[c]{@{}c@{}}1. Deep Convolutional+SVM: 93.35\%±0.62\\      2. Deep Fully Connected+SVM: 94.44\%±1.21\\      3. CNN: 95.35\%±0.42\end{tabular} \\ \cline{2-3} 
 &
  \textbf{Proposed Method} &
  \textbf{\begin{tabular}[c]{@{}c@{}}Accuracy: 97.87\%\\ Precision: 98.56\%\\ Recall: 99.12\%\\ F1-score: 98.89\%\end{tabular}} \\ \hline
\end{tabular}
}
\end{table}

%
\subsection{McNemar's Statistical Test}
\textcolor{black}{The McNemar's statistical test has been performed in the present work, for the statistical analysis of the proposed classification framework. For this, the proposed model has been compared to the CNN models from which the features were extracted and used for the final classification. The results are shown in Table \ref{mcnemars}. To reject the null hypothesis that the two models are similar, the $p-value$ from the McNemar's test should remain below 5\% (i.e., 0.05), and from the table, it can be seen that for every comparison case, the $p-value<0.05$. Thus, the null hypothesis can be rejected and it can be concluded that the proposed model is dissimilar to any of the feature extractor models and performs superior to them. Thus statistical analysis of the proposed model justifies the reliability of the approach devised in this research.}

\begin{table}[]
\centering
\caption{\textcolor{black}{Results obtained from McNemar's statistical test. For all three datasets, the proposed framework is compared to the CNN models whose features have been used. The $p-value$ is less than 0.05 for every case and thus, the null hypothesis is rejected.}}
\label{mcnemars}
\begin{tabular}{|c|c|c|c|}
\hline
\textbf{McNemar's   Test} & \multicolumn{3}{c|}{\textbf{p-value}}                                                 \\ \hline
\textbf{Performed with}   & \textbf{Herlev   Pap Smear} & \textbf{Mendeley   LBC} & \textbf{SIPaKMeD   Pap Smear} \\ \hline
ResNet-50    & 0.0046 & 0.0012 & 0.0005 \\ \hline
VGG-16       & 0.0001 & 0.0211 & 0.0007 \\ \hline
DenseNet-121 & 0.0103 & 0.0089 & 0.0315 \\ \hline
Inception v3 & 0.0007 & 0.0061 & 0.0100 \\ \hline
\end{tabular}
\end{table}

\section{Conclusions and Future Work}\label{conclusions}
The need for automation in the cervical cancer detection domain arises due to the high mortality rate throughout the close. Motivated by this cause, we developed a fully automated detection framework that optimizes deep features for classification. The two-level enhancement boosted the classification performance while simultaneously reducing the training time significantly. This research also explored the hybridization of multiple CNN-based deep features to extract more discriminating information from the dataset.

An alternative way of feature selection is exalted in this research that uses Principal Component Analysis (PCA) and Grey Wolf Optimization (GWO). The two-level feature reduction approach introduced in this paper leverages the advantages of both methods resulting in optimal feature set selection. The proposed method achieves better results juxtaposed to end-to-end classification with CNN models, while simultaneously reducing the computation cost. Very high classification accuracy of 99.47\%, 98.32\%, and 97.87\% on the three publicly available benchmark datasets, namely Mendeley LBC, Herlev Pap Smear and SIPaKMeD Pap Smear datasets respectively tantamount to state-of-the-art methods.

However, there is scope for further improvement by utilizing different classification models and using hybrid metaheuristic feature selection algorithms. This paper craved a path for further research in this field as well as multi-domain adaptation. \textcolor{black}{The proposed pipeline can be used as a test-bed for several classification problems, not only in biomedical applications but in other computer vision problems as well. The feature selection can be further addressed by developing an end-to-end multi-objective hybrid optimization algorithm, that selects optimal feature set, where the objective function aims to increase the classification performance by selecting the least number of features, thereby reducing the computational cost simultaneously. }

\section*{Acknowledgements}
The work is supported by SERB (DST), Govt. of India (Ref. no. EEQ/2018/000963).

\section*{Conflict of interest}
The authors declare that they have no conflict of interest.

\bibliographystyle{spbasic}
\bibliography{References}

\end{document}